# MobileDepth: Efficient Monocular Depth Prediction on Mobile Devices

Yekai Wang
*School Of Electronic Engineering And Computer Science*
*Faculty of Science and Engineering*
*Queen Mary, University of London*
London, United Kingdom
y.wang@se19.qmul.ac.uk

*Abstract*—Depth prediction is fundamental for many useful applications on computer vision and robotic systems. On mobile phones, the performance of some useful applications such as augmented reality, autofocus and so on could be enhanced by accurate depth prediction. In this work, an efficient fully convolutional network architecture for depth prediction has been proposed, which uses RegNetY 06 as the encoder and split-concatenate shuffle blocks as decoder. At the same time, an appropriate combination of data augmentation, hyper-parameters and loss functions to efficiently train the lightweight network has been provided. Also, an Android application has been developed which can load CNN models to predict depth map by the monocular images captured from the mobile camera and evaluate the average latency and frame per second of the models. As a result, the network achieves 82.7% δ1 accuracy on NYU Depth v2 dataset and at the same time, have only 62ms latency on ARM A76 CPUs so that it can predict the depth map from the mobile camera in real-time.

## I. Introduction

Depth prediction is a basic task in computer vision which can be beneficial to other applications in both robotic systems (e.g. obstacle detection, 3D mapping, etc.) and computer version (e.g. semantic segmentation, object detection, etc.). It is worth researching deploying depth prediction on mobile systems because there are some useful applications could be significantly enhanced by depth prediction, e.g. augmented reality, auto focus, etc. Depth map could be generated from scenes with both hardware and software. Hardware-based depth prediction is based on distance sensors which are expensive and space-consuming (e.g. 3D laser scanners, cameras with structured light, etc.) so that they can hardly be installed on mobile systems. In software-based methodologies, with the rapid development on deep learning, recent works (Eigen et al 2014, Liu et al 2015, Laina et al 2016, Zhang et al 2018, Wu et al 2019, etc.) have proved that fully convolutional neural networks (Long et al 2015) could reconstruct depth maps from monocular RGB images in high performance.

However, most of the depth prediction works are based on complexed CNNs which can be hardly run on ARM mobile CPUs on real-time. To solve this problem, the efficient networks proposed in the recent years pretrained on ImageNet (Deng et al 2009) could be used as the encoder to extract the high-level depth features from the input monocular RGB images, and the convolutional blocks proposed from popular CNNs could be used as the decoder to reconstruct the depth map from the extracted features. A fully convolutional neural network (Long et al 2015) with lightweight encoder and decoders would be able to be run efficiently on mobile CPUs in real-time.

In this work, several experiments have been done to find an appropriate combination of dataset augmentation, loss functions, encoders, decoders and skip connection to build a CNN network which can be efficiently run on mobile ARM CPUs with high performance on evaluating metrics. As a result, the network proposed in this paper achieves 82.7% δ1 accuracy (defined in section III-D) which is 5.6% higher than the prior work (Wofk et al 2019) on efficient depth prediction in similar latency on ARM CPUs (the comparison is shown in Fig. 1). Also, a benchmark tool on Android has been developed in this work which can predict the depth map from the mobile phone camera in real-time and at the same time can evaluate the average forward speed and frame per second of the model.

Fig. 1. Comparison with prior work: (a) the input RGB image
(b) ground truth (c) Wofk et al's (2019) network, unpruned, δ1=77.1%, latency=55ms (d) this work, δ1=82.7%, latency=62ms
The color indicates the depth which is mentioned in the color bar

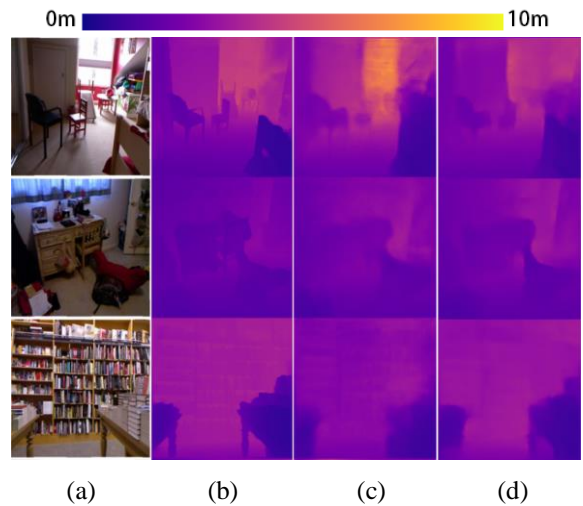

## II. Related Work

In this section, several related researches done in the past on supervised monocular depth prediction, lightweight convolutional neural networks and mobile phone implementation.

### A. Monocular Depth Prediction

Reconstructing depth maps from single RGB images is a worth-study research topic on computer vision because traditional 3D reconstruction and recognition methods need extra hardware (for example, stereo camera, laser scanner, IR sensor and so on) and cost much time in the process, however, monocular depth prediction from single images based on CNNs not only shows promising performance, but also have low hardware cost and fast prediction speed. Based on the idea that both global structure and local clues are needed to predict depth from monocular images, Eigen et al (2014) designed a system consists with two networks: one coarse network makes rough prediction from the entire image and the other fine network optimizes the prediction results locally. The evaluation metrics on the prediction performance are widely used in the future works. Liu et al (2015) proposed an fully convolutional network (Long et al 2015) based model to learn the unary and pairwise energy of continuous conditional random field in a unified network and a novel superpixel pooling approach to improved the speed. Laina et al (2016) proposed a fully convolutional network (Long et al 2015) with residual blocks and firstly applied reversed Huber loss function on depth estimation task. Zhang et al (2018) proposed a novel task-attention module so that one single network could have two parts of blocks focusing on semantic segmentation task and depth estimation task respectively by recursive learning. Xu et al (2018) proposed a CRF-based (Liu et al 2015) model where structured attention blocks are implemented between the multi-scale outputs of the encoder and the CRF module so as to automatically learn the importance of the features. Wu et al (2019) proposed a generative adversarial network (Goodfellow et al 2014) based depth estimation model from monocular video serial with attention mechanism which can automatically learn the importance of each channels. Wofk et al (2019) proposed a lightweight encoder-decoder network MobileNet v1-NNConv5 with additive skip connection (the encoder is MobileNet v1 and the decoder is NNConv5 which has 5 depthwise separable convolutional layers as the hidden layers and a pointwise convolutional layer as the output layer combined with 5 nearest-neighbor upsample interpolation functions to upsample the output layer becoming the same size as the input) which not only can be efficiently run on embedded systems but also achieves 77.1% $\delta 1$ accuracy on NYU Depth v2 dataset (Silberman et al 2012), and the network MobileNet v1-NNConv5 is set as the baseline of this work.

*B. Lightweight Convolutional Neural Networks*

Howard et al (2017) proposed an efficient network (MobileNet v1) with depthwise separable convolutions (i.e. a depthwise convolution on each channel followed by a pointwise convolution to learn linear combination of the channels) to get similar accuracy and at the same time save over 88% FLOPs in comparison to full convolutional networks. Also, MobileNet v1 can be scaled by a hyper-parameter to gain appropriate trade-off between performance and efficiency. Sandler et al (2018) proposed MobileNet v2 with inverted residuals and linear bottlenecks (i.e. firstly expand the channels by pointwise convolution, then filter with depthwise convolution, finally project back by pointwise convolution without activation function, also, a short cut is applied for residual computation). In MobileNet v3 (Howard et al 2019) and EfficientNet (Tan and Le 2019), the activation functions in the blocks are replaced with swish and Se Modules (i.e. squeeze-and-excitation modules) (Hu et al 2018) are implemented to automatically learn channel-wise importance of the blocks (i.e. attention mechanism). Also, neural architecture search technology (Zoph and Le 2016) is used to search the network by compounding width scaling, depth scaling and resolution scaling. Radosavovic et al (2020) proposed a new exemplification for designing design spaces of neural architecture search and as a result, got simple and regular RegNet which can be run up to 5 times faster rather than EfficientNet on GPUs with similar accuracy. In this work, the lightweight CNNs mentioned above will be used as the encoder of the FCNs (Long et al 2015) for depth prediction task.

*C. Mobile Phone Deployment*

It has become more and more important to efficiently deploy neural networks on edge devices such as mobile phones because artificial intelligence is becoming useful in mobile applications (e.g. speech recognition, object classification, etc.). One generic way for mobile deployment is to convert the network model to a generic model by open neural network exchange system (ONNX 2017) at first and then use TVM runtime (Chen et al 2018) or other frameworks to make the model runnable on mobile phone, however, this method is complicated and error prone because the standard of the operators and network structures are quite different among the frameworks. Torch (Paszke et al 2017) released a mobile version called torch mobile in 2019 which allows keeping within torch ecosystem from training to deployment, which makes the deploy process easier and more stable. In this work, torch mobile will be used to deploy the depth prediction network on mobile ARM CPUs.

Fig. 2. Proposed network architecture in this work. The blue blocks represent the encoders of RegNetY 06 (Radosavovic et al 2020) which extracts the high-level features from the RGB image at resolution of 224x224 to 7x7 with 608 channels. The decoders (defined in section III-C Fig.3) reconstructs the depth map from the high-level features of 608 channels. The output layer is a nearest neighbor interpolation followed by a pointwise convolutional layer from 32 channels to a single depth channel (for better visual effect, the depth map here is converted to a coloured map). The blue arrows represents the additive skip connection to transmit the information in the encoders to the decoders for spatial feature reuse to reconstruct the depth map with more clear edge and shape.

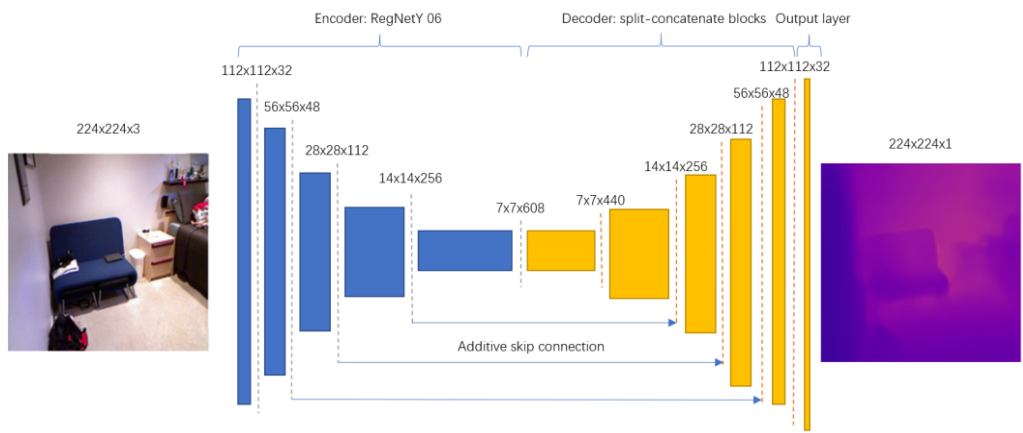

## III. METHODOLOGY

In this section, the approach to the final result of the project will be provided, including requirements analysis, model architecture design, mobile application implementation and performance evaluation.

### A. Requirements Analysis

This project aims at developing an efficient application on mobile phones with limited computing resources to predict the depth from monocular cameras as accurate as possible. This application should at first allow users to capture image stream from their mobile phones' cameras. Then, an efficient lightweight convolutional neural network which can output the depth maps from the inputted RGB images should be deployed in the application. After that, the application should be able to output the depth maps from the image stream in real time. Finally, this application should provide performance benchmark, for example, frame per second, time per image and so on to prove that it can run in real time.

According to the requirements above, the key to the success of the project is to develop an efficient and fast convolutional neural network which can be run on mobile phone CPUs. Therefore, this project will put emphasis on designing an efficient CNN to predict depth maps from monocular images.

### B. Network Architecture

As is shown in Fig. 2, this work proposed a fully convolutional network (Long et al 2015) using RegNetY 06 as the encoder and split-concatenate blocks as the decoder with additive skip connection between encoders and decoders so that the decoders can get high-level features. The structure of the split-concatenate blocks, which is inspired from ShuffleNet v2 (Ma et al 2018), is shown in Fig. 3. With such blocks, the latency could be reduced by 3ms in comparison to normal depthwise separable convolutional blocks and at the same time, gain higher performance on evaluation metrics.

Fig. 3. Structure of split-concatenate block. The input is at first upsampled with nearest neighbor interpolation. Then, the channels are split to two. The first part is put into a pointwise convolutional layer and the second part is put into a with a pointwise convolutional layer and a depthwise separable convolutional layer. The output of the two parts are concatenated to get the output channels. After that, the channels are shuffled to build cross talk among the features. At last, an encoder layer is additively connected to the output

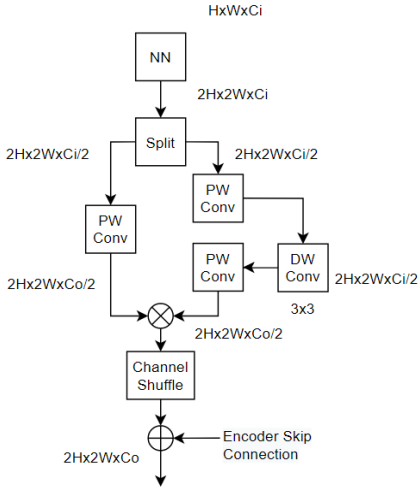

### C. Mobile benchmark application implementation

Fig. 4 shows the benchmark tool developed on Android and implemented torch mobile framework (Paszke et al 2017) to run the trained models. In this work, the tool is run on ARM A76 CPU @2.6GHz to evaluate the model forward time on mobile phones, which is an important metric in ablation study.

Fig. 4. Mobile benchmark tool. (a) shows the main interface of the benchmark tool. The top-left box displays the input from the camera and the box below shows the predicted depth (darker is closed and vice versa). The text below shows the frame per second, the current model forward latency and the average latency in 300 forward times. The "LOAD MODEL" button enables users to choose a model file for depth prediction from the storage. "USE CAMERA" button will enable the camera and predict the depth in real-time. After clicking "BENCHMARK" button, the camera will be disabled and zero tensors will be inputted into the model to calculate the average model forward duration because different images from real scene may effect the model forward speed. (b) is the file explorer to load models.

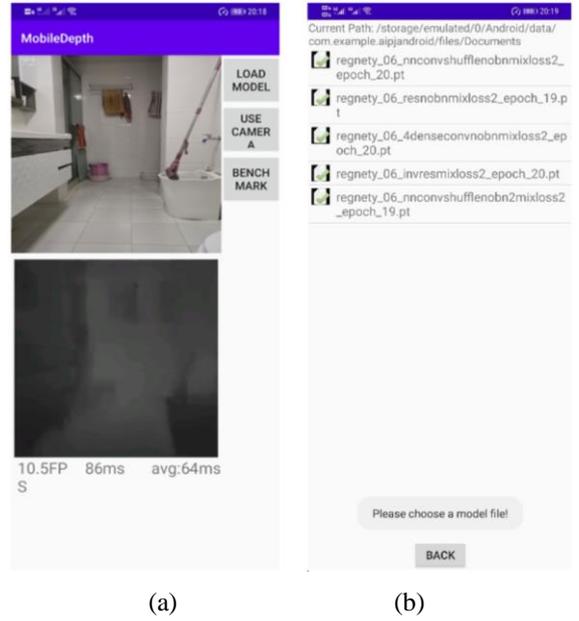

(a)          (b)

### D. Evaluation

The quality of the models are compared and evaluated by the performance metrics used in the previous works:

- Root Mean Squared Error(RMSE):
$$RMSE = \sqrt{\frac{1}{n}\sum_{i}^{n}(y_i - \hat{y}_i)^2} \quad (1)$$

- Average Relative Error(REL):
$$REL = \frac{1}{n}\sum_{i}^{n}\frac{|y_i - \hat{y}_i|}{y} \quad (2)$$

- Accuracy under Threshold($\delta_i$):

$$\% \text{ of } y_i \text{ s.t.} \max\left(\frac{\hat{y}_i}{y_i}, \frac{y_i}{\hat{y}_i}\right) = \delta < T, T = 1.25, 1.25^2, 1.25^3 \quad (3)$$

To evaluate the speed of the models, the latency metrics are used including number of parameters, MAdds (multiply-accumulate operations) and average forward duration time by 300 times benchmarked on ARM A76 CPU @2.6GHz (the mobile phone is physically cooled down after each benchmark to avoid the influence of CPU frequency reduction caused by high temperature).

From Table I, this work has better performance than some of the prior works on complicated network. Compared with

Wofk et al's (2019) work, this work has far better performance with a little bit higher latency.

TABLE I. COMPARISON AMONG PRIOR WORKS AND THIS WORK

| Network | Evaluation Metrics | | | | |
|---|---|---|---|---|---|
| | MACs [G] | RMSE | $\delta 1$ | Rel | CPU [ms] |
| Eigen and Fergus (2015) | 20.40 | 0.641 | 0.769 | 0.158 | 578 |
| Laina et al (2016) | 32.30 | 0.573 | 0.811 | **0.127** | 685 |
| DORN (Fu et al 2018) | 68.17 | 0.509 | 0.828 | 0.115 | - |
| Wofk et al (2019), unpruned | 0.74 | 0.599 | 0.775 | - | **55** |
| This work | **0.70** | **0.497** | **0.827** | 0.137 | 62 |

## IV. EXPERIMENTAL RESULTS

In this section, the experimental results as well as the comparison to the existing works will be provided. At first, the details of the dataset and the implementation will be described. Then, the ablation study result will be provided comparing among loss function design space, encoder, decoder, skip connection and attention mechanism.

### A. Dataset

The networks and their accuracy are trained and evaluated on NYU Depth v2 (Silberman et al 2012) which contains both RGB images and depth maps taken by Microsoft Kinect from 464 different indoor scenes at resolution of 640x480. The models are trained on a subset of 50K training set provided by Alhashim and Wonka (2018) (inpainting method (Levin et al 2004) is used to fill the missing values) and tested on 654 official split testing set at resolution of 224x224. During testing process, the average of the prediction of the original images and their mirror images are evaluated.

### B. Implementation Details

The depth prediction models are implemented in PyTorch (Paszke et al 2017) and trained with 32-bit floating point precision on one NVIDIA RTX 2070 super GPU with 8GB memory. The Adam optimizer is used for network optimization. The training batch size is set to 16, the learning rate is initialized by 0.0002 and betas are set to (0.9, 0.999) (Alhashim and Wonka 2018). Training is performed for 20 epochs on each model with pretrained parameters on ImageNet (Deng et al 2009) loaded to encoders and Kaimin normal (He et al 2015) weight initialized to the decoders.

### C. Ablation Studies: Data Augmentation

Data augmentation during training process could have positive effect on reducing the overfitting on the training set. In this work, random color channel swap and random horizontal flip is implemented for data augmentation (Alhashim and Wonka 2018).

Also, by iterating over the NYU Depth v2 dataset (Silberman et al 2012), it is found that the distribution of the depth is imbalanced that the values concentrates upon small range of nearer depth, but limited on further depth (shown in Fig. 5). This kind of distribution may lead the network to have more attention on nearer depth but not robust to further depth after training process. To deal with this problem, the model trained by the inversed or log ground truth may be more robust to the depth prediction objective because it can show the decreasing confidence on the increasing depth (Ummenhofer et al 2017).

In the experiment on NYU Depth v2 training dataset, the depth ground truth is at first divided by 255 to be normalized between 0 and 1. Then, it is clamped between 0.01 and 1 to avoid infinite number to get $y_{normal}$. Finally, the inversed ($y_{inversed}=1/y_{normal}$) and log ($y_{log}=\ln(y_{normal})$) ground truth are used to train the model. In this part, the model is fixed with MobileNet v1-NNConv5 with skip add connection (Wofk et al 2019). As a result of this experiment shown in Table II, the inversed ground truth has much better performance on this task as its $\delta 1$ is 3.54% higher than the normal ground truth. The log ground truth is hard to converge so that it is not provided in this table. Therefore, the inversed ground truth is chosen in the later experiments.

Fig. 5. Depth distribution over normalized NYU Depth v2 dataset. The frequency of normalized depth from 0 to 1 increases quickly at first and drops rapidly after the peak point. The depth from 0.5 to 1 account for a very small part of the data. Training on such imbalanced data would make it difficult to learn larger values.

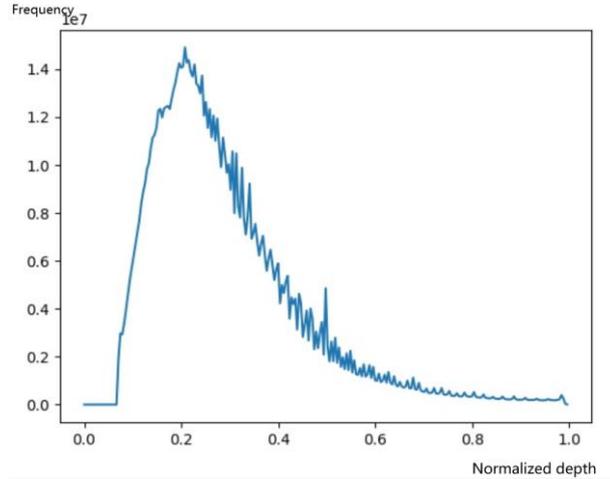

TABLE II. COMPARISON BETWEEN NORMAL AND INVERSED GT

| Ground Truth | Evaluation Metrics | | | | | |
|---|---|---|---|---|---|---|
| | RMSE | MAE | REL | $\delta 1$ | $\delta 2$ | $\delta 3$ |
| Not Inversed | 0.592 | 0.440 | 0.168 | 0.762 | 0.938 | 0.982 |
| Inversed | 0.552 | 0.404 | 0.153 | 0.790 | 0.949 | 0.986 |

### D. Ablation Studies: Loss Function

An appropriate loss function can significantly reduce the converge speed during training process and make the final results of the models perform better. In depth estimation tasks, L1 loss function (4) and berHu loss function (5) are commonly used. L1 loss function indicates pixel-wise difference between the depth maps:

$$L_{L1} = \frac{1}{n}\sum_{i}^{n}|y_i - \hat{y}_i| \qquad (4)$$

BerHu loss function is a reversed Huber loss function (Owen 2007) which shows a trade-off between L1 loss and L2 loss with a threshold c. The threshold is computed from 20% of the maximum error in one batch:

$$L_{BerHu} = \frac{1}{n}\sum_{i}^{n}|y_i - \hat{y}_i|_b, |x|_b = \begin{cases} |x| & |x| \leq c \\ \frac{x^2+c^2}{2c} & |x| \leq c \end{cases} \qquad (5)$$

However, the loss functions above could only measure the pixel-wise error but could hardly measure the image structure similarity. Image gradient loss can compute the error of the reconstruction of the edge to make the edge in the depth map sharper and more clear (Alhashim and Wonka 2018). The edge loss function is defined as:

$$L_{Edge} = \frac{1}{n}\sum_{i}^{n}|g_x(y_i, \hat{y}_i)| + |g_y(y_i, \hat{y}_i)| \qquad (6)$$

Structural similarity (Wang 2004) can measure the performance of image reconstruction so that it could have good effect on depth map generation. The maximum of SSIM is 1 when the two images are same on the structure so the loss function is defined as:

$$L_{SSIM} = \frac{1-SSIM(y,\hat{y})}{2} \quad (7)$$

It has been proposed in the previous works that appearance matching loss (Godard et al 2017) which combines with SSIM loss function and L1 loss function could well indicate the photometric reconstruction performance on depth prediction tasks. The loss is defined in (8), where α (set to 0.85 here) is the weight between the two components:

$$L_{AM} = \alpha L_{SSIM} + (1-\alpha) L_{L1} \quad (8)$$

Alhashim et al (2018) proposed a loss function (9) combining L1 loss with edge loss and SSIM loss with weight ω (it is set to $\omega_1=1.0, \omega_2=1.0, \omega_3=0.1$ here to penalize L1 loss function) so that it could not only minimize the pixel-wise error but also reduce the distortion of the image structure in the reconstructed depth map. The loss function is defined as:

$$L_{MIX} = \omega_1 L_{SSIM} + \omega_2 L_{Edge} + \omega_3 L_{L1} \quad (9)$$

In this experiment, the model is fixed as MobileNet v1-NNConv5 with skip add connection (Wofk et al 2019). As a result, $L_{BerHu}$ performs better than $L_{L1}$ in all the metrics. The appearance matching loss shows similar performance as the BerHu loss. The mix loss has worst performance than $L_{BerHu}$ and $L_{AM}$, but much better than $L_{L1}$. From the result, using $L_{BerHu}$ instead of $L_{L1}$ in $L_{MIX}$ and $L_{AM}$ may have better performance on the task. Experiments have been done to validate this guess by defining the new loss function as:

$$L_{BMIX} = \omega_1 L_{SSIM} + \omega_2 L_{Edge} + \omega_3 L_{BerHu} \quad (10)$$

During the experiment, it is found that assigning a too small weight (i.e. $\omega_3=0.1$) to BerHu loss could lead to worse performance. By observing the components of the $L_{BMIX}$, the BerHu loss exceeds the $L_{SSIM}$ and $L_{Edge}$ an order of magnitude so it may be better to set a dominant weight to $L_{BerHu}$ to gain fast convergence and let $L_{SSIM}$ and $L_{Edge}$ to support to reconstruct the depth map similar in image structure. From the result shown in Table III, the model performs best on RMSE, MAE, REL and δ1 when the three components have the same weight in $L_{BMIX}$, therefore, the experiments hereafter will fix the loss function as $L_{BMIX}(10), \omega_1=1, \omega_2=1, \omega_3=1$.

TABLE III. COMPARISON BETWEEN L1 LOSS AND MIX LOSS

| Loss Function | Evaluation Metrics | | | | | |
|---|---|---|---|---|---|---|
| | RMSE | MAE | REL | δ1 | δ2 | δ3 |
| $L_{L1}$ (4) | 0.552 | 0.404 | 0.153 | 0.790 | 0.949 | 0.986 |
| $L_{BerHu}$ (5) | 0.537 | 0.394 | **0.144** | 0.802 | **0.957** | **0.989** |
| $L_{AM}$ (8), α=0.85 | 0.535 | 0.390 | **0.144** | 0.805 | 0.956 | **0.989** |
| $L_{MIX}$(9), $\omega_1=1, \omega_2=1, \omega_3=0.1$ | 0.532 | 0.392 | 0.147 | 0.800 | 0.955 | 0.988 |
| $L_{BMIX}(10), \omega_1=1, \omega_2=0, \omega_3=0.1$ | 0.545 | 0.405 | 0.153 | 0.789 | 0.950 | 0.987 |
| $L_{BMIX}(10), \omega_1=1, \omega_2=1, \omega_3=0.1$ | 0.540 | 0.397 | 0.152 | 0.795 | 0.951 | 0.987 |
| $L_{BMIX}(10), \omega_1=1, \omega_2=0, \omega_3=1$ | 0.545 | 0.400 | 0.150 | 0.797 | 0.954 | 0.988 |
| $L_{BMIX}(10), \omega_1=1, \omega_2=1, \omega_3=1$ | **0.531** | **0.389** | **0.144** | **0.807** | 0.954 | 0.988 |

*E. Ablation Studies: Encoders*

In this part, the model MobileNet v1-NNConv5 proposed by Wofk et al (2019) is used as the baseline. The decoder is fixed as NNConv5 (Wofk et al 2019) and the encoders are altered among several lightweight networks including MobileNet v1 (Howard et al 2017), v2 (Sandler et al 2018), v3 (Howard et al 2019), EfficientNet(L0, L1 and L2) (Tan and Le 2019) and RegNetY(02, 04, 06 and 08) (Radosavovic et al 2020). The final blocks including the final average pooling layers and the output fc layers are removed and connected with the decoder to become fully convolutional networks (Long et al 2015). In this part, skip connection is at first not used for the consideration of that the output channels between the encoders and the decoders should be matched when using skip connection, which may effect the fairness of the experiment.

TABLE IV. COMPARISON AMONG ENCODER DESIGN SPACE

| Encoder | Evaluation Metrics | | | | | |
|---|---|---|---|---|---|---|
| | Weights [M] | MACs [G] | RMSE | δ1 | Rel | CPU [ms] |
| MobileNet v1 | 4.99 | 0.76 | 0.550 | 0.791 | 0.152 | 55 |
| *With skip add* | *4.99* | *0.76* | *0.531* | *0.807* | *0.145* | *54* |
| MobileNet v2 | 4.40 | 0.52 | 0.551 | 0.795 | 0.154 | 53 |
| *With skip cat* | *4.42* | *0.55* | *0.529* | *0.808* | *0.143* | *60* |
| MobileNet v3 | 6.20 | 0.40 | 0.560 | 0.787 | 0.154 | **50** |
| *With skip cat* | *6.22* | *0.44* | *0.537* | *0.798* | *0.149* | *54* |
| EfficientNet L0 | 5.54 | 0.57 | 0.552 | 0.795 | 0.152 | 58 |
| *With skip cat* | *5.62* | *0.62* | *0.529* | *0.809* | *0.143* | *65* |
| EfficientNet L1 | 6.31 | 0.69 | 0.552 | 0.796 | 0.152 | 68 |
| *With skip cat* | *6.38* | *0.74* | *0.521* | *0.816* | ***0.139*** | *73* |
| EfficientNet L2 | 6.98 | 0.77 | 0.555 | **0.797** | **0.146** | 74 |
| *With skip cat* | *7.06* | *0.82* | *0.520* | *0.819* | *0.140* | *79* |
| RegNetY 04 | 4.78 | 0.57 | 0.575 | 0.772 | 0.163 | 68 |
| *With skip add* | *4.61* | *0.51* | *0.529* | *0.813* | *0.143* | *66* |
| RegNetY 06 | 6.58 | 0.77 | 0.548 | 0.794 | 0.154 | 69 |
| *With skip cat* | *6.51* | *0.74* | ***0.503*** | *0.820* | *0.142* | *68* |
| RegNetY 08 | 6.88 | 0.97 | **0.539** | **0.797** | 0.150 | 78 |
| *With skip add* | *6.92* | *0.99* | *0.510* | *0.821* | *0.140* | *79* |
| *With skip cat* | *6.99* | *1.11* | ***0.503*** | *0.825* | *0.141* | *95* |

From the results without using skip connection shown in Table IV, MobileNet v1, MobileNet v2, EfficientNet L0, L1, EfficientNet L2, RegNetY 06 and RegNetY 08 have similar performance while MobileNet v3 and RegNetY 04 have relatively worse performance. In general, too small encoders (e.g. MobileNet v3 and RegNetY 04) performs worse but the networks with more parameters and calculations cannot provide proportional better performance. This may because of that the encoder-decoder architecture cannot reuse the low-level features in the encoder without skip connection so that the image details are lost when the encoder extracting the high-level features in low resolution. The skip connection can help the network to reconstruct the depth map with sharper edge and more similar image structure by merging high resolution features in the encoder. By experimenting both concatenative and additive skip connection on RegNetY 08-NNConv5, it is found that concatenative skip connection need more computing resource, but have little profit on the accuracy and performance. Inverted residual block with bottleneck which expands the channel inside the residual block are used in MobileNet v2, v3 and EfficientNet L0, L1, L2 so that the output channel number of each block is low. If use additive skip connection on these encoders, the channel number of the decoder will be reduced too much to match with the encoder channels, which may decrease the accuracy of the decoders. However, using concatenative skip connection on these encoders may slightly increase the computation cost by approximately 6%-10% due to their low output channel of the blocks. Therefore, concatenative skip connection is implemented on these MobileNet v2, v3 and EfficientNet L0, L1, L2 and additive skip connection is implemented on the other encoders.

From the result shown in Table IV, RegNetY 08 with concatenative skip connection achieves the best performance but cost much higher computing resource. MobileNet v3 with concatenative skip connection has the lowest latency but has unsatisfactory performance, this may because that MobileNet v3 is searched by NAS (Zoph and Le 2016) to have best trade-off between latency and performance on ImageNet (Deng et al 2009) classification task, but its architecture may be unable to adopt to depth prediction task via transfer learning. To better display the data, the latency and δ1 accuracy of the networks are normalized and put in one chart (Fig. 6). The points nearer to the top-left side of the chart show better performance and lower latency, so, it is obvious that RegNetY 06 with additive skip connection has the best trade-off between latency and accuracy. In the later parts, RegNetY 06 is picked as the main research object.

Fig. 6. Normalized Lantency-Accuracy Chart from the data in Table IV. The points nearer to the left have lower latency and the points nearer to the top have higher accuracy. RegNetY 06 is the point nearest to top-left so that it has the best trade-off between latency and accuracy.

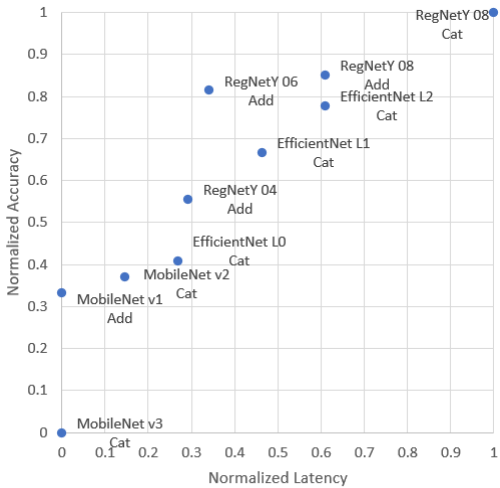

### F. Ablation Studies: Decoders

In this part, the encoder is fixed as RegNetY 06 and the decoder is researched and designed in several aspects: with or without batch normalization, interpolation mode, decoder structure and different activation functions.

*1) Batch normalization:* Sergey and Christian (2015) proposed batch normalization to stabilize and speed up the training process of deep networks by normalizing channel-wise parameters in each mini-batch, which has been widely used in deep CNN models. However, Lim et al (2017) proposed that the information of image scale in super resolution tasks will be lost after applying batch normalization and removing batch normalization can help saving up to 40% memory cost to implement deeper models. In FCNs (Long et al 2015), the decoders can be considered as super-resolution problems. An experiment has been taken in comparison of RegNetY 06-NNConv with or without batch normalization layers in the decoder. The training loss and the δ1 accuracy on the testing set during training process is shown in Fig. 7. The evaluation result of network with or without batch normalization is shown in Table V.

Fig. 7. Training loss and testing accuracy chart w/wo batch normalization. The model without batch normalization converges much faster and has less computational cost and higher performance.

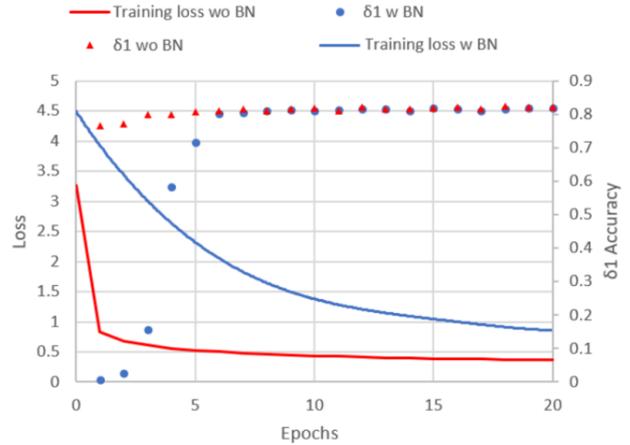

TABLE V. COMPARISON ON MODELS W/WO BATCH NORMALIZATION

| Decoder | Evaluation Metrics | | | | | |
|---|---|---|---|---|---|---|
| | *RMSE* | *δ1* | *δ2* | *δ3* | *Rel* | *CPU [ms]* |
| With BN | 0.503 | 0.820 | 0.961 | **0.990** | 0.142 | 68 |
| Without BN | **0.500** | **0.824** | **0.962** | 0.989 | **0.139** | **65** |

As a result, the model without batch normalization converges much faster and have better performance on evaluation metrics. The experiments hereafter will remove all batch normalization layers on the decoders.

*2) Interpolation mode:* to restore the depth map from low resolution feature maps extracted by the encoders, bilinear interpolation and nearest neighbour interpolation are considered between the decoder convolutional blocks to build skip connection with the encoder in multi-scales because these two algorithms are fast. The result of the experiments are shown in Table VI. Bilinear has a little better performance than NN interpolation, however, it costs approximately 15% more latency on mobile CPUs. In consideration of the trade-off between efficiency and performance, nearest neighbour interpolation is used as the upsampling function hereafter.

TABLE VI. COMPARISON ON MODELS WITH DIFFERENT INTERPOLATION MODE

| Interpolation | Evaluation Metrics | | | | | |
|---|---|---|---|---|---|---|
| | *RMSE* | *δ1* | *δ2* | *δ3* | *Rel* | *CPU [ms]* |
| NN | 0.500 | 0.824 | 0.962 | 0.989 | **0.139** | **65** |
| Bilinear | **0.498** | **0.826** | **0.963** | **0.990** | **0.139** | 75 |

*3) Skip connection number and output layer:* The skip connection from different layers of the encoder may effect the feature level the decoders can reuse. Inspired from a super-resolution CNN (Lin et al 2020), additive skip connection could be built from the image input to the depth map output with a pointwise convolutional block to gain global residual learning.

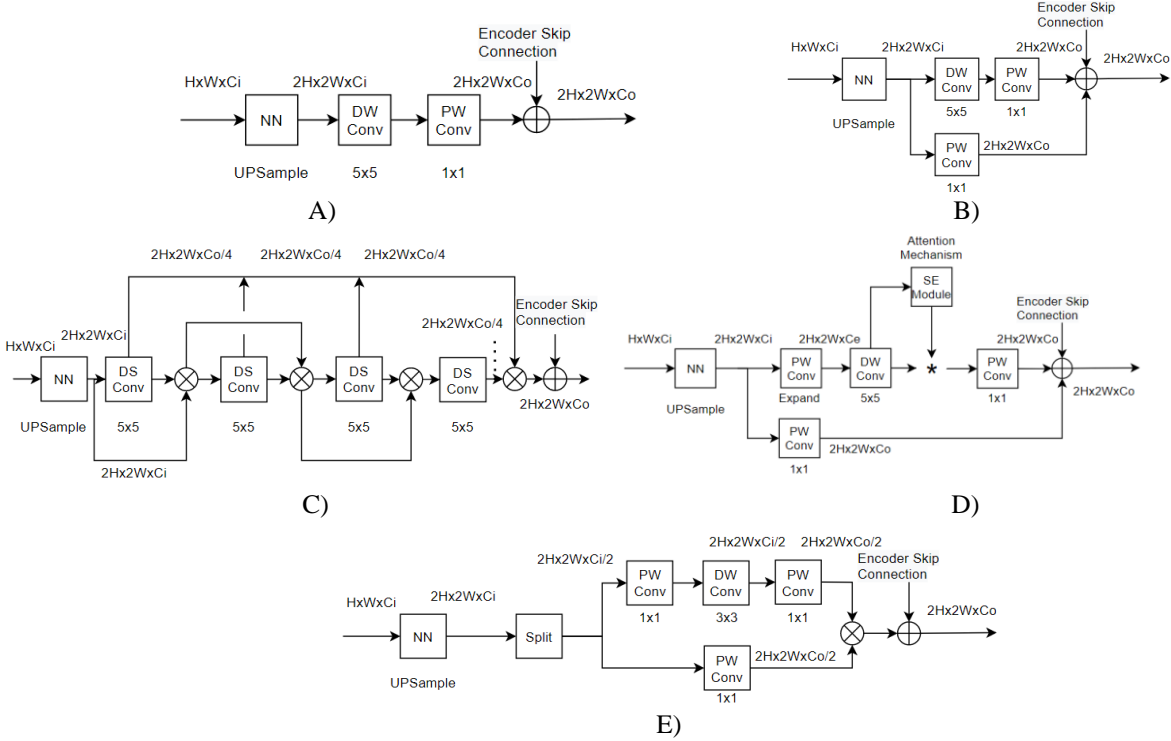

Fig. 8. Decoder Structure Diagrams. The symbol "+" means additive connection, "X" means concatenative connection, "NN" means nearest neighbor interpolation, "Ci" means input channel, "Co" means output channel, "Split" means equally split the channels of the layer to two, "SeModule" is "Squeeze-and-Excitation" block (Hu et al 2018), the detail is shown in Fig. 9, "DS Conv" means depthwise separable convolution, "DW Conv" means depthwise convolution and "PW Conv" means pointwise convolution. (a) is the depthwise separable convolutional block; (b) is the residual depthwise separable convolutional block; (c) is the dense concatenated depthwise separable convolutional block; (d) is the inverted residual and linear bottleneck with SE module (Howard et al 2019); (e) is split-concatenate shuffle block (Ma et al 2018).

TABLE VII. COMPARISON ON SKIP CONNECTION NUMBER

| Skip connection | Evaluation Metrics | | | | | CPU [ms] |
|---|---|---|---|---|---|---|
| | RMSE | $\delta 1$ | $\delta 2$ | $\delta 3$ | Rel | |
| Encoder-Decoder(2,3,4) | **0.500** | **0.824** | 0.962 | 0.989 | 0.139 | **65** |
| Encoder-Decoder(2,3,4,5) | 0.504 | **0.824** | 0.962 | **0.990** | **0.138** | 66 |
| Encoder-Decoder(2,3,4,5) &Input-Output | 0.504 | 0.823 | **0.964** | **0.990** | 0.139 | 74 |

TABLE VIII. COMPARISON ON OUTPUT LAYER

| Output Layer | Evaluation Metrics | | | | | CPU [ms] |
|---|---|---|---|---|---|---|
| | RMSE | $\delta 1$ | $\delta 2$ | $\delta 3$ | Rel | |
| Pointwise | 0.500 | 0.824 | 0.962 | 0.989 | 0.139 | **65** |
| Depthwise separate | **0.496** | **0.826** | **0.963** | **0.990** | **0.138** | 78 |

From Table VII, the performances are similar which may because that the important features are extracted in later stages in the encoder but not the initial block nor the input image so that in depth prediction task, these two skip connections could hardly have effect, which is different from super-resolution task. Therefore, 3-skip connection of lowest latency is chosen.

Also, a single pointwise convolutional block was used in the final output layer. By observing the depth map, it would be better to replace the single pointwise output layer with a depthwise separable convolutional layer to get smoother and more clear output.

As is shown in Table VIII, the depthwise separate output layer has a little bit better performance than the single pointwise layer, however, it takes much higher latency because the output resolution is too high which could take high computational cost. Therefore, pointwise output layer is chosen.

*4) Decoder Structure:* Inspired by serveral popular network architectures, 5 lightweight decoder structures are considered in this part. In the experiments, five decoder blocks are cascaded and additive connection is applied between encoders and decoders as mentioned in part 3. The decoder blocks are finally linked to an output layer to predict depth maps. Followed by the structures of the decoder blocks:

Depthwise separable convolutional block: a normal depthwise separable convolutional block with 5x5 kernel size (Fig. 8 a)

Residual depthwise separable convolutional block: this block is inspired by ResNet (He et al 2016) in which the depthwise separable learns the residual information (Fig. 8 b)

Dense depthwise separable convolutional block: this one is inspired by DenseNet (Huang et al 2017). This decoder block has four densely concatenated depthwise separable layers which have 1/4 output channels and the layers are finally concatenated together to gain full output channel (Fig. 8 c). This structure may help reused of the features of the former layers.

Inverted residual and linear bottleneck block: this block is the structure from MobileNet v3 (Howard et al 2019), which has inverted bottleneck and residual learning. Also, squeeze-and-excitation block (Hu et al 2018) is implemented to automatically learn the attention on each channel (Fig. 8 d).

Split-concatenate shuffle block: this block is inspired from ShuffleNet v2 (Ma et al 2018). The input channels are at first split to two and sent to one pointwise convolutional layer and one depthwise separable convolutional layer respectively. Then, the two branches are concatenated to get output features. At last, the features are shuffled so that cross talk could be built among the features (Fig. 8 e).

TABLE IX. COMPARISON AMONG DIFFERENT DECODER STRUCTURES

| Decoders | Evaluation Metrics | | | | | |
|---|---|---|---|---|---|---|
| | Weights [M] | MACs [G] | RMSE | $\delta 1$ | Rel | CPU [ms] |
| DS (A) | 6.51 | 0.73 | 0.500 | 0.824 | 0.139 | 65 |
| ResDS (B) | 6.92 | 0.83 | 0.505 | 0.822 | 0.136 | 74 |
| DenseDS (C) | 6.75 | 0.89 | 0.503 | 0.822 | 0.138 | 100 |
| InvRes (D) | 8.34 | 1.10 | 0.514 | 0.821 | **0.135** | 103 |
| SplitShuffle (E) | **6.43** | **0.70** | **0.497** | **0.827** | 0.137 | **62** |

From the result shown in Table IX, decoder E has the best performance on almost all the metrics as well as the lowest latency. The complicated decoders do not provide increasement on the performance, which may because that the encoder RegNetY 06 is quite efficient so that its precision is not high enough to match more complicated decoders. Therefore, the split-concatenate shuffle block is chosen as the decoder.

*4) Attention mechanism:* Convolutional blocks combine spatial and channel features together in local fields to extract information, however, the importance of the channels and space may be different in the network layers, so, it could be effective to increase the performance of the network if the model can automatically learn the importance information. Hu et al (2018) proposed a novel unit: "Squeeze-and-Excitation" SE block (shown in Fig. 9) which can be easily applied on different CNN blocks and can be efficiently run with low computational cost.

Fig. 9. The structure of "Squeeze-and-Excitation" block (Hu et al 2018). The blocks above learns the importance weights of the channels and the weights are multiplied to the channels.

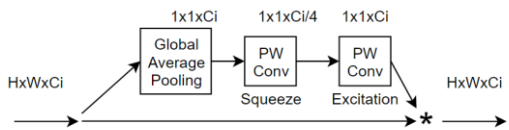

In this experiment, the channel-wise attention mechanism is implemented on the blocks of the decoders. From the result shown in Table X, the network with or without SE block has similar performance, which may because that in such a small decoder (split-concatenate shuffle block), the additional SE blocks may have little effect when the encoder RegNetY 06 has attention mechanism already.

TABLE X. COMPARISON AMONG DIFFERENT DECODER STRUCTURES

| Attention Mechanism | Evaluation Metrics | | | | | |
|---|---|---|---|---|---|---|
| | Weights [M] | MACs [G] | RMSE | $\delta 1$ | Rel | CPU [ms] |
| Without SE | **6.43** | **0.70** | **0.497** | **0.827** | 0.137 | **62** |
| With SE | 6.61 | 0.71 | 0.500 | 0.826 | **0.135** | 70 |

## V. FUTURE WORK

In this work, the lightweight network is trained with simple supervised learning. Some better training method may be effective on the depth prediction task (e.g. self-training with noisy student (Xie et al 2020) to increase the performance and generalization of the model, etc.). For the network architecture, a more efficient decoder with appropriate skip connection and attention mechanism could be searched by reinforcement learning (Zoph and Le 2016) rather than hand-designed models if enough computational resource is available. Also, network prune may be effective to remove the redundant neural cells to significantly reduce the model latency. From another angle, other network architecture could be explored because perhaps there is a better way to reconstruct the depth map rather than traditional fully convolutional networks.

## VI. CONCLUSION

This work has provided a combination of data augmentation, loss functions, encoders, decoders and skip connection and got an efficient network with low latency and high accuracy to enable real-time depth prediction on mobile systems with high confidence. Also, a mobile depth prediction benchmark tool has been developed in this project to evaluate the model latency.

## ACKNOWLEDGMENT

I would like to thank to my supervisor for his criticism, guidance and support during the process of this project. I would like to thank to the Queen Mary University of London for the excellent academic environment and resources.